# Moral Dilemmas for Artificial Intelligence: a position paper on an application of Compositional Quantum Cognition

Camilo M. Signorelli[1,2,5][0000-0002-2110-7646], Xerxes D. Arsiwalla[3,4,5]

[1] Department of Computer Science, University of Oxford, Oxford, United Kingdom
[2] Cognitive Neuroimaging Lab, INSERM U992, NeuroSpin, Gif-sur-Yvette, France.
[3] Barcelona Institue of Science and Technology, Barcelona, Spain
[4] Institute for BioEngineering of Catalonia, Barcelona, Spain
[5] Pompeu Fabra University, Barcelona, Spain
cam.signorelli@cs.ox.ac.uk

**Abstract.** Traditionally, the way one evaluates the performance of an Artificial Intelligence (AI) system is via a comparison to human performance in specific tasks, treating humans as a reference for high-level cognition. However, these comparisons leave out important features of human intelligence: the capability to transfer knowledge and take complex decisions based on emotional and rational reasoning. These decisions are influenced by current inferences as well as prior experiences, making the decision process strongly subjective and "apparently" biased. In this context, a definition of compositional intelligence is necessary to incorporate these features in future AI tests. Here, a concrete implementation of this will be suggested, using recent developments in quantum cognition, natural language and compositional meaning of sentences, thanks to categorical compositional models of meaning.

## 1 Introduction

Moral dilemmas and a general intelligence definition have been recently suggested as an alternative to current AI tests [1]. Usually, Intelligence is interpreted regarding particular and efficient behaviours which can be measured in terms of performing or not these behaviours. One example is the Turing test [2], in fact, the first approach grounded on human behaviour and the most well-known and controversial test for AI. Other examples are challenging machines in games like chess or Go [3], and testing AI programs with dilemmas [4], theoretically, demanding a more complex level of information processing. Nevertheless, all these approaches lack a general definition of intelligence as a minimal requirement to measure intelligence [5–7], and restrict AI only to humankind "intelligence" without including key features of a true human intelligence. Therefore, this position paper is going to shortly introduce a new strategy and research program to quantify a probably more complete and inclusive evaluation for future advances in AI.

The discussion will start with a preliminary definition of general intelligence; some key ingredients in human intelligence will be analyzed to finally arrive at more general principles of compositional intelligence. Then, a Moral test emerges naturally, as an option to effectively identify a combination of different and important features in human intelligence. Additionally, the issue about how to implement a Moral test is sketched, using new advances in the categorical compositional model of meaning (CCMM) in natural language [8] and the emergent field of Quantum Cognition (QC), associated with cognition and decision making under uncertainty [9].

## 2 Towards a General Definition of Intelligence

### 2.1 What is Intelligence?

A useful way to approach this question is by conceptualizing general intelligence as the capability of any system to take advantage of its environment in order to achieve a specific or general goal [1]. Biologically speaking this goal would be surviving, or in reductive terms: trying to keep the autonomy and potential reproduction of the system; while the goal in machines can be solving a specific task or problem using internal and external resources. This advantage would take place as a balance of these resources or in other words, reducing disequilibrium between them. This balance is managed internally, as the cognitive architecture/process to deal with internal and external demands, using both internal and external feedback. If any animal, human or machine break the balance between internal and external resources, as for example achieving their particular needs at the expense of the total annihilation of their environment and resources, this animal, human or machine is leading its own annihilation, which is not particularly intelligent. Therefore, a general intelligence is defined here as the balance (reducing disequilibrium) between these external and internal resources like an embodied system [10]. This general definition can incorporate living beings as well as robots and computers, and in this way, intelligence is general enough to include different kind of intelligence, multiple intelligence, contextual influences and different kind of systems with different degrees of intelligence [1]. Balance, as a "relative efficiency", is considered as part of a general feature of intelligent systems and it can be linked with a physical view of intelligence using entropy or complexity [6, 7, 11, 12]. The mathematical description of these concepts is expected in future developments.

### 2.2 Human Intelligence and Quantum Cognition

A preliminary definition of human intelligence can be suggested as the ability to benefit or gain advantage from their social environment while maintaining autonomy[1, 13]. This requires a balance or equilibrium between rational and emotional information processes [1, 10]. Hence, humans would solve problems by trying to incorporate different types of information and balancing both internal and external resources.

However, common approaches on human intelligence assume only a rational and logical nature of human thinking. Against that, many different cognitive experiments have been shown how human thinking can be "easily wrong" or "illogical", what is associated with some cognitive fallacies [14]. Cognitive fallacies are typically wrong assumptions about the dynamics of cognition and judgments. Experimental examples are usually wrong answers to apparently simple questions, which contradict classical probabilistic frameworks [15]. These answers are apparently due to fast and intuitive processing before some extra level of information processing, evidencing that the dynamic for holistic information processing can be completely different from logical or rational processing of the same information. These cognitive features show interesting properties of concept combinations, human judgments and decision making under uncertainty. For example, one of these cognitive fallacies is the ordering question effect: the idea that the order in which questions are presented should not affect the final outcome or response. However, in psychology and sociology, this effect is well known and called question bias. For example, analyses in 70 surveys from 651 to 3006 participants demonstrated not only that question order affects the final outcome, but also this effect can be predicted by quantum models of cognition, revealing a non-commutative structure [16]. Another example is the conjunction of two concepts like the Pet-fish problem, where the intersection of individual concept probabilities do not explain the observed probability for a typical Pet-Fish like a goldfish [17]. Recently, the explanation of these phenomena were demonstrated using the mathematical framework of quantum mechanics; also known as the quantum cognition (QC) programme [9]. QC uses non-classical mathematical probabilities to successfully explain and predict part of these cognitive behaviours. One of these predictions is the recently proved constructive effect of affective evaluations [18], where preliminary ratings of negative adverts influence the rating of following positive adverts and vice versa. It suggests that the construction of some new affective content has intrinsic connection with QC formalism. In consequence, understanding human intelligence is not possible with only the classical idea of logical and rational kind of intelligence, it is also needed to explain and incorporate emotional and intuitive intelligence (no classical reasoning), apparently better described by QC.

In this way, we will suggest a compositional human intelligence (Fig 1a) composed of a rational reasoning and also intuitive and emotional one, considering intelligence as the whole "living" body engaged with the environment [10] ("brain-body-environment" system). Compositional complexes intelligences would be different in the way how they manage both (or even more) rational and emotional processes of information associated with internal and external resources.

### 2.3 In search of the General Principles of Intelligence

The order-effect of stimuli presentations, conjunction, disjunction, decision making under uncertainty, contextuality, among others, can be understood as an intrinsic property of human judgment: contextual dependencies as inherent to previous knowledge experiences. In other words, humans and animals can understand/distinguish among different contexts and act accordingly, thanks to previously learned experiences. Therefore, one reasonable, but not an intuitive assumption, is to consider these effects as a general property (or the minimal requirement) of high-level cognition, i.e. some types of bias would be a reaction to contextual dependencies, as wording or framing, implicit meanings, question order, etc. For instance, exchanging words like "kill" instead of "save" in moral judgment, triggered different answers even if the outcomes of each dilemma were the same in both situations [19]. If some changes in the formulation of dilemmas trigger different contextual internal meanings and they evoke different answers according to different contexts, it would mean that high-level cognition can understand and distinguish different context and respond differently to each one in a way that is not always optimal (rationally speaking).

A true understanding of different contexts (in the sense of [20]) implies the existence of a minimal kind of meaning. This understanding of context due to meaning (e.g. BUP does not mean the same than PUB) would require a minimal non-commutative structure of cognition connected with our suggested principle of balance internal and external resources. We hypothesized this structure as part of an inherent neural network construction in the brain, where structure-function relationship would be dynamic, highly flexible and context-dependent [10]. Emotion and rational reasoning, in this sense, would correspond to the outcome of interactions among many components, each one related to neural assemblies or distributed networks, which combine, influence, shape and constrain one another [1, 21].

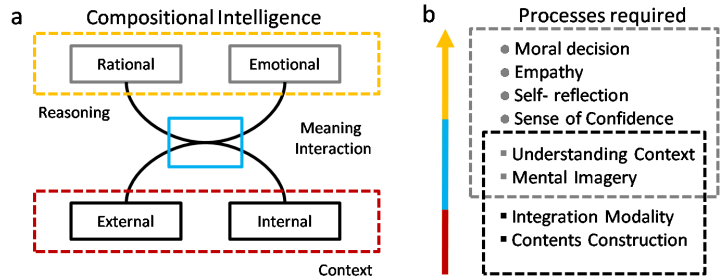

**Fig 1**. a) A proposal for Compositional Intelligence. Context is constructed by composition of external and internal objects; the interaction evokes and/or creates meaning from which intuitive/emotional and rational reasoning would emerge. b) Moral Test and Processes required. Some processes needed for moral thought are stated as examples, among many other possible processes.

To sum up, the idea of balancing internal and external resources as a preliminary proposal for a general intelligence, implies at least four requirements (Fig 1): i) a non-commutative structure (mathematically and operationally), ii) understanding of context, which implies different kind of contexts effects, framing, wording, question order-effects, etc, iii) meaning projection in at least emotional and rational meaning spaces, which implies a basic notion of subjectivity (see discussion on section 4), and iv) behaviours, judgments and decisions based on these meanings (reasoning would emerge from these meaning interactions).

## 3      Towards a Moral Test

Even if these previous requirements were defined regarding human cognition, it is expected that any kind of true intelligence can understand different contexts and evoke, create or recreate meaning of their internal and external resources. Particularly, in humans, our definition of human intelligence implies that humans can integrate different types of information and manage a balance between internal and external social resources through intuitive, emotional and rational reasoning, which, in turn, would emerge from meaning interactions. Thus, the next question is how to measure and quantify these requirements, both in humans and machines.

One way to answer this question is searching for situations where humans need to explicitly use both emotional and rational resources to solve complex problems. One example is a certain kind of moral dilemmas. Moral dilemmas are controversial situations to study moral principles, where subjects need to judge some actions and sometimes take difficult and even paradoxical decisions. Moral dilemmas are simple, in the sense that they do not require any kind of specific knowledge, but at the same time, some of them can be very complex because they require a deep understanding of each situation, and deep reflection to balance moral consequences, emotions and optimal solutions. No answer is completely correct, they are context dependent and solutions can vary among cultures, subjects, or even across the same subject in particular emotional circumstances. Moreover, morality and ethics are not necessarily associated with a particular religion, political view, education level, age or gender [22, 23], while it seems an intrinsic human condition and a very "relative" (or even biased) feature. People that give the impression to act against the moral establishment, really act according to their "own" moral, apparently developed in completely different social accepted conditions. One simple example is the acceptation of monogamy or polygamy and therefore certain moral attitudes in different societies. Morality, in this way, is the set of internal (particular experience) and external (social culture) values learned by experience, which allows us to behave in our societies on demand of both emotional and rational thoughts, usually reacting in a very intuitive and fast way. Morality also requires many previous processes associated with high-level cognition (Fig 1b), starting for decision-making to self-reflection, to be able to detect mistakes on these decisions; sense of confidence, to estimate how correct a decision or action is; mental imagery, to create new probable scenarios of action; empathy, to equilibrate individual and social requirements; understanding of context, to adapt moral decisions to the context, among others. Therefore, the main suggestion is that moral and ethics emerge as a way to integrate individual and social regulation (as different types of information) in human species, and apparently also in other animals species [24] (even if human moral can be completely different in comparison with animal moral). Morality is related to both rational and emotional reasoning [25] and it has the peculiarity to be very dependent of the context, wording and framing [19, 23], kind of social community, subjects and probably even emotional states of each subject [26].

Hence, any precise test to measure the distinct human intelligence (or even non-anthropomorphic intelligence) should consider the way of thinking and information processing to develop moral thoughts, independently of what is judged as a correct or incorrect about these thoughts in our societies. It is how the dynamics of answers and meanings to moral dilemmas change depending on the context, the capacity to justify or not any action, and report our intrinsic experience according to intuitive and rational reasoning, what is really important in human intelligence and what should be measured as an attribute of intelligence (balance between rational and emotional processes).

## 4    Compositional Quantum Cognition

In order to implement the idea of a Moral test, it is necessary a compositional cognitive model and specifically a model of natural language able to incorporate the non-commutative, bias and commutative properties at different levels and different contexts of natural answers to different moral dilemmas (where rational and emotional thoughts are involved). For one side QC seems a good candidate for cognitive model while CCMM is for natural language. CCMM has been proved a better theoretical framework than only distributional and symbolic models of natural language [27] and more general than some QC models [28]. Even though some of these experiments have been only made in statics text corpus, they suggest a potential richer description for more complex cognitive experiments.

Therefore, in this section and in order to maintain simplicity, the main concepts of CCMM framework will be presented together with a preliminary framework, which expects to integrate QC and CCMM in a common approach. For a more complete mathematical background and description about CCMM and QC, please refer to [8, 29] and [9, 15, 30] respectively.

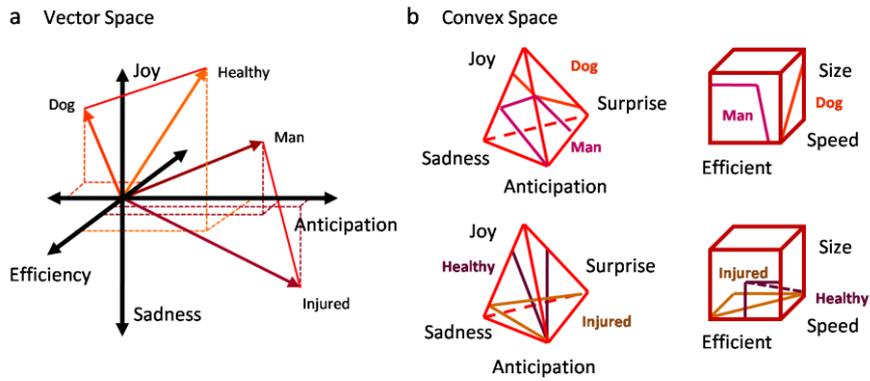

**Fig 2**. Semantic space with emotional and rational basis. a) One example of Vector Space for words: Dog, Man, Healthy and Injured. The basis is a combination of emotional (Joy vs. Sadness) and rational categorization (Efficient vs. Inefficient). b) Example of Convex Space for the same words above. One emotional convex space (Joy, Sadness, Anticipation, Surprise) and other rational convex space (Big, Fast, Efficiency). In convex spaces, words correspond to regions of the space, instead of only vectors as it is in vector space.

### 4.1 Categorical Compositional Model of Meaning

Applications of CCMM basically needs: 1) Define a compositional structure as a grammar category (e.g. pre-groups grammar); 2) Define a meaning space as a semantic category, for example vector space or conceptual space of meaning; and finally, 3) Connect both categories in some way that it is possible to interpret the grammar category into the semantic category (mathematically, one has to define a functor between categories) [8]. Thus, some useful concepts are:

**Compositional Structures.** In CCMM, compositional structures are certain rules/definitions about how elements compound each other. In other words, how processes, states, effects, among other possible elements, compound. Grammatical types and their composition are described using a pregroup algebra due to Lambek [31]. However, any kind of grammar definition can, in principle, be implemented. Grammar will be interpreted as the way how word meanings interact, defining first primitive types as nouns $n$, sentences $s$, and then other types like adjectives $nn^l$, and verbs, for instance, a transitive verb as $n^r s n^l$, among other kinds of words to form complex sentences.

**Semantic Spaces.** Semantic spaces are spaces where individual words are defined with respect to each other. The simplest way is using distributional approaches to define vectors of meaning for each word (Fig. 2a), or even better, defining density matrices. The choice of a basis vector and how to build other words, adjectives and verbs from the basis, is not trivial and it can be done in many different ways. Of course, it will depend on what the experimenter would like to describe and compare.

**Conceptual Spaces.** The idea of conceptual spaces, recently suggested in [32], is a more cognitively realistic way to define semantic spaces. This approach is called convex conceptual spaces. In short, concepts can be defined by a combination of others primitive features or quality dimensions, building spaces which can be superposed or not, to define regions of similarity. One perceptual example is to define taste based on some features such as: Saline, Sweet, Sour, and Bitter. Then, different kind of food would be described with a certain level of each taste dimension, and where other features like colour, texture, can also be incorporated [29]. Other complex example is defining elements regarding emotional states and factual features (Fig. 2b). Additionally, conceptual spaces require two semantic/meaning spaces, one for words in a quality dimension space (Fig 2) and other for sentences in a "sentence meaning space" (Fig 3a), then, the final sentence meaning is an "interaction" between both spaces.

**Computing the meaning of a sentence.** Diagrammatically, the final meaning of one sentence will be the meaning of individual words interacting according to the grammatical structure, defined as a process [8, 29].

### 4.2 Proof of Concept: Compositional Quantum Cognition

In our framework, meaning is the interaction between external and internal objects, understanding external objects/contents as transductions of external stimuli, while internal objects/contents would correspond to the "space of transduction" that will help to create internally, these external objects[1]. These objects can have different levels of complexity; some of them can act as a constitutive element in the construction of other contents or being formed by other more fundamental elements. Specifically, internal objects can create different levels of what we call "quality dimensions" (as a basis space) and external objects are formed by and move in these quality dimensional spaces. In other words, external objects are defined with respect to internal objects which will form a type of internal space depending on the specific problem to model.

Regarding our specific hypotheses, these internal objects and/or quality dimensions can be emotions (usually based on belief) or reasons (usually based on facts), while external objects will be a convex space (for each object) defined by these emotions and reasons (Fig 2b). Both, internal spaces and external objects create what in other frameworks is called "state of mind", usually represented by Ψ. However, in our approach, it is not a simple state; Ψ will correspond to a complex internal space and ex-ternal objects which are embodied into another "decision space" (Fig 3a). This decision space could be a similar space as defined in QC. This is not; however, the end of the story, this decision space is again embodied into a bigger "meaning space" (sentence space), where decisions/behaviours/thoughts/judgments/concepts among others, can project into meanings (Fig 3b). The sentence meaning space is created by other level of internal objects, in this case "internal values", so any decision/behaviour/thought/judgment/concept would have a meaning corresponding to different superposition of values (Fig 3c). Thus, in this framework, meaning will be the projection/interaction of convex external objects into a first layer of internal object space which through a decision/behaviour/thought/judgment/concept project another meaning in a second layer of internal object space (thanks to reasoning). Since each person would have a unique internal space in each layer (even if they share the same axes, the topology can be different), the meaning is intrinsically related to subjectivity. In this sense, different contexts would trigger different or similar meanings (projections/interactions) and in consequence decision/behaviour/thought/judgment/concept would have different or similar meanings, depending on the type of interaction.

---

[1] We avoid the term representation because in the literature it has been invoked with many different connotations.

If the interaction corresponds to a rational process, the meaning would be relatively fixed across different situations, but if it is emotional/intuitive, the meaning will change with the situation (unfixed meaning). Additionally, the beliefs in the emotive component and the facts on the rational part of an external object can be known or unknown by the subject. If they are known, the subject can access and report these beliefs or facts, in some way that question (in the "decision space") about them only fill a third-person lack of information, however when they are not known, the subject needs to create or re-create them in order to report internally (first-person) and/or externally (third-person) something about them. In that situation, elements about the potential report would be also part of the creation, re-creation or co-creation of these beliefs or facts. As suggested in [15, 33, 34] and others, classical probability (CP) would mainly apply when subjects known about their contents (classical-deterministic reasoning) while quantum probability (QP) will work better when they are unknown (non-classical-determinist reasoning, or intuitive reasoning).

Thus, both descriptions would be part of, at least, two kind of intelligence, that all together and in a compositional way, would converge to a more complete description of human intelligence, something that we preliminary call a compositional approach of human cognition or compositional intelligence and it is expected to be deeper developed in future works as a compositional quantum cognitive approach or even a more general compositional contextual model of cognition.

Summarizing, our framework is a complex three space interaction, where external objects can be created by emotional and rational quality dimensions (internal objects), which can be known or unknown by the subject (following CP or QP respectively), having an impact in the evocation, creation or co-creation of fixed or unfixed meanings, which in turn expose emotional/intuitive or rational reasoning (Fig 1a and Fig 3).

## 5      Implementation of a Moral Test

Now, we have all the elements, at least conceptually, to describe a research program of implementing a moral test. This strategy has one theoretical and three experimental research steps. The main goal in our implementation is keeping the simplicity of the written exchange of questions and answers and analysing them using non-commutative models of language at different levels, looking for bias meaning and "subjectivity" from previous knowledge.

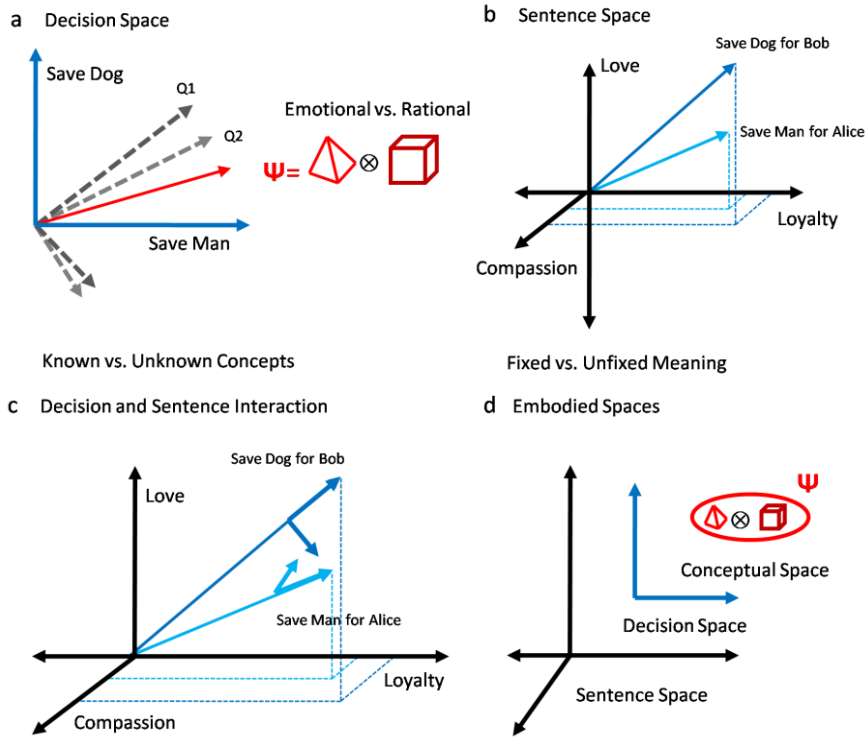

**Fig 3**. a) Decision Space would correspond to the usual QC description. These decisions can be previously known or unknown by the subject. Ψ, the psychological state, is defined based on emotional and rational quality dimensions. b) Sentence Space is a second level of meaning definition (for sentences). In this example the basis is defined regarding some values: Compassion, Love and Loyalty. Sentence meaning can be fixed (static) or unfixed (dynamical). c) Decision and sentence/meaning interaction. Decisions would be explained in a sentence meaning space. In this example, two different decisions, one for Alice (save the man) and other for Bob (save the dog), hidden slightly similar meanings. d) The big picture: Quality dimensions would be embodied into the other complex decision and meaning spaces.

## 5.1 Step I. Mathematical model definition

Moral dilemmas seem better than simple day to day questions, especially to test some of the requirements and hypotheses stated above. For instance, manipulating contexts in the same moral dilemma will allow us to measure the sequential change in responses and how they follow a classical probability combination when only fixed thinking is present, non-classical probability combination in only intuitive or unfixed thinking and finally how the dynamic of responses changed when the dilemma confronts both emotional and rational thoughts. For example, in experiments were subject were asked to justify their moral judgments, most of them failed to properly explain their choices, revealing high intuitive (emotional) components, instead of rational ones [22] and suit-able to be described by QC more than classical frameworks. Hence, we suggest that if the subject rationalizes the dilemma, or situations are manipulated to strongly confront emotional (intuitive) versus rational thoughts, the outcome/meaning would change, and the decision or judgment will be more difficult.

In order to do that, the kind of moral dilemmas required in our implementation should confront emotional and rational resources to judge actions, in a way where utilitarian decisions should not be intuitive, while the probable human answer would be predictable, and only after reasoning (emotional or rational), the non-predictable option would emerge as a possibility. So, our set of moral dilemmas and paradigm need to be slightly different from the existing ones in the current literature.

According to the intelligence definition above, purely emotional or rational judgments would not be intelligent decisions in these contexts; instead, an intelligent decision would balance both possibilities until reaching the best compromise for each subject (in his/her space of meaning). It means that measuring the probability of each answer (as usually in QC) is not enough; it is also needed to compute the meaning behind of each answer. This can be done asking directly to the subject why he/she took one or another decision (asking for justification), what is closely related to a phenomenological approach [35] and computing their answers thanks a natural language model.

Thus, a more complex mathematical model is needed, both to define external and internal objects and the interactions to compute and compare meanings across different subjects, sentences and even machines. It requires the developing of both a novel experimental paradigm (variations of experiment in Appendix A) and adaptation of CCMM into an ideally more complete compositional model of cognition, what was conceptually defined in section 4.2 and where mathematical definitions are expected as previous requirement to next steps.

## 5.2   Step II: A Toy Example, Moral dilemmas and context effects

A simple example of experiment would have a structure like this: 1) Definition of a conceptual semantic space on quality dimensions and space of meaning. It means to define a set of internal objects (quality dimensions) like emotions and reasons (facts) which will correspond to the axes, and the projections of "external" concepts into these axes, forming a convex space for each concept (Fig 2). 2) Then, a set of different versions of the same dilemma where only a few concepts are changed to confront different degrees of emotional and rational values. 3) Each dilemma would have one introduction to the story and context, one question associated with the agreement of two or more different actions (Question 1), another question associated with the personal decision (Question 2) and a final "why" question to justify their choice (Question 3). Question 1 and Question 2 can be exchanged to show order effects and/or constructive affective effects. One example is in Appendix A.

The moral dilemma in Appendix A is a variant with three different questions and can be manipulated in many different ways to confront emotional and rational thoughts. So, first, a set of different versions of the same moral dilemma would allow us to describe the understanding of each context and perhaps even looking for contextuality, and secondly, measuring the time reaction for each version will be a way to quantify the dynamic of each judgment (easy, difficult, fast, and slow). If our hypotheses are correct, it is expected that the answers will depend of the modifications in the dilemma formulation and degree of conflict among dilemmas (rational vs. emotional).

Other variations of these ideas and maybe a better way to test the QC phenomenon can be incorporated in the final experiment, for example, using conjunction, disjunction, among other effects observed in QC. Additionally, after each moral dilemma, we can ask for the degree of emotional arousal using standard cognitive tests.

### 5.3  Step III. Quantification of meaning in Moral dilemmas

The conceptual convex space defined above should be a conceptual subjective space from which the meaning of crucial words (external objects) in each dilemma will be previously defined according to the individual categorization of words in a fixed or flexible quality dimensional space (internal objects) of at least two layers.

It can be done in many different ways. For example, some emotional and rational quality dimensions can be defined as internal objects and each subject would be able to determine concepts (external objects) according to different values of these quality dimensions (Fig 2 and 3). Each subject will be asked to evaluate (in a scale of points) some words with respect to these dimensions, building a convex space for each crucial word and subject. Other words can be defined thanks affective lexicon from [36]. Another option is to directly ask the meaning of each crucial word, for instance: What does a dog mean to you? Or what does a man mean to you? Etc. Consequently, the answer can be taken as the direct meaning of each concept and use it to reduce sentences. However, this second option is less simple than the first option.

With the semantic space for each subject, the answer to the "why" question can be quantified in terms of the sentence meaning (interaction or projection into word and sentence space) for each sentence. These answers and sentences contain the essential features of a moral thought and the meaning is expected to change from simple meaning to more complexes, depending on how complex is the dilemma (more or less degree of emotional and rational confrontation).

One consequence of this analysis would be the possibility to predict or at least correlate the final decision according to the subjective meaning of individual words, with respect to each dilemma. At the same time, general meanings can be inferred even when decisions can be completely different. For example, if "Bob likes Dogs", it is possible that Bob would save the dog instead of the man, while if Alice, who hypothetically does not like dogs, would likely do the opposite. However, if a quality dimension is defined with respect to emotions vs. reasons, and sentence meaning with respect to values like love, loyalty among others (Fig 3), the final meaning for Bob saving the Dog could be similar than the meaning of saving the man for Alice (Fig 3b). In other words, apparently different decisions would have similar meanings, and the opposite can be also true: same decisions hide completely different ones.

The way how these meanings evolve from context to context is what we refer here as one general property of high-level cognition associated with QC effects and subjectivity, which in the end can be quantified and compared thanks CCMM. In other words, QC and CCMM will be used to compute some kind of subjectivity in a way that is compared across subjects, making this approach a novel tool to complement the phenomenological program suggested by Varela in [35].

### 5.4  Step IV. Application for AI

The last step would be the implementation of moral dilemmas in AI programs using QC to search for non-commutative structures and CCMM to compute their "why" answers. Thanks to some variations in CCMM and QC (step 1), the decisions and meaning of answers would be directly computed and the structure of meaning across human compared with the meaning across different instances of AI programs, developing a way to compare subjective features across humans and machines.

This implementation is not something trivial and will require a big effort in both, previous validation of moral dilemmas in human (preliminary steps) and then adaptations for AI. For example, it is expected to arrange similar experiences than step II, and simulate different instances to compute the same effect (if there is or not) comparing changes in the kind of structure for different versions of our dilemmas. Concretely, AI would be adapted to answer the same experimental set-ups for human, but any modification in order to facilitate or not the answers of the software will be avoided. Then, QC effects will be quantified in the same way than humans and the meaning of their answers will be "computed" using the same strategy than in human experiments, from the semantic space defined by the machine.

Therefore, the first attempt would be searching for a simple "understanding" of context, i.e. if AI programs can distinguish and differentiate versions of the same moral dilemma, with slightly different words and how the answers change in comparison with the observed changes in humans (framing, wording, question order effects, affective construction, etc). The second attempt is looking for meaning structure exploring the "why" answers of AI and the connection with its own semantic spaces. Specifically, in the moral test, the biggest effects are expected when the moral dilemma confronts emotional and rational thoughts. One hypothesis is that current AI will not be able to answer these questions, but if they can, even if answers and meanings could be similar, the structure and dynamics of their meaning will be different from the structure of meaning in humans across dilemmas. It can be quantified and compared thanks to the experimental probabilistic distributions of answers in both humans and machines.

## 6     Discussion and Conclusion

It is reasonable to expect that even if machines can demonstrate completely different logical or "illogical" answers, the general way of thinking dealing with different contexts in moral dilemmas should be more or less generic among species, including machines, if they reach high-level cognition. In other words, contextual dependency, QC effects, meaning and subjectivity built on previous and inferred knowledge should be captured by a complete model of cognition, which would be able to identify and measure these features. Thus, interference, conjunction and disjunction, wording, question ordering effects, and distributed meaning could be expected as general features of rational and emotional thoughts altogether, and where specific dynamics would emerge confronting some kind of moral dilemmas.

Hence, one suggestion of this position paper is that a moral test can help quantify these differences in a particular semantic space characterized with both emotional and rational components. Thus, a machine would reach part of what is defined as compositional human intelligence if the machine is able to show autonomously speaking (in the sense of defining their own goals), the intricate type of thinking that humans have when they are confronted with these kinds of dilemmas, even if their answers can be completely different from ours, the structure and dynamics of meaning is expected to be similar. In other words, the possibility or not to first understand/differentiate con-

4text, second project external objects into internal objects to have meaning and third argue any decision based on rational and emotional components, is what is intrinsically related to a complex individual and social intelligence, which characterize humans and should be expected in some high-level cognitive AI. If subjects can justify their choice, it would imply that emotion and reason were playing some kind of role, while if they fail to justify their actions, only intuitive processes were involved. Can we expect something similar in the current AI approaches?

Finally, these views imply another ethical problem regarding the kind of morality which would emerge in AI. For example, independently of how one implements morality in AI (please see [1] for restrictions), there are important concerns about replicating morality in AI and using moral tests: if the replication of human moral process (dynamics and QC effects) in AI is desirable, this replication can be dangerous since AI is different from human, so AI will also develop a different kind of morality, based on certain type of subjectivity; other way around, if the exact replication of this moral process is not desirable, why moral dilemmas, QC effects and cognitive fallacies would be a useful tool? First, the Moral test suggested here will try to evaluate the understanding of context (through QC effects), evolution of meaning (through CCMM) and the effects of confronting rational and emotional thoughts (through changing the degree of rational and emotional components among versions of each dilemma). In this sense, morality itself (as purely set of rules) is not necessarily desirable (especially the biased morality), but the balance among rational and emotional thoughts is what we claim can be really desirable as a high-level compositional intelligence. Then, we argued, morality emerges from these interactions. Moreover, our framework here was following the anthropomorphic approach, it implies that AI is compared with human moral dilemmas. This is a contradictory strategy regarding our general intelligence case, however, with that we expect to show the paradoxical and ethical consequences of the anthropomorphic views [1, 37]. In this sense, the risk of dangerous machines is exactly the same risk of dangerous humans, and before worrying about dangerous AI, we should first care about making a psychological healthy world, and in consequence humans and machines would share similar social behaviours. Of course, a truly non-anthropomorphic test should consider the same two first elements (context and meaning), but the balance can be through other types of reasoning and depends on the autonomous specification of machine goals.

## Appendix A: Example of a Moral dilemma

After a shipwreck, a **healthy dog** and an **injured man** are floating and trying to swim to survive. If you are in the emergency boat with only one space left:

Please indicate your degree of agreement with the next options (where +5 strongly agree, +3 moderately agree, +1 slightly agree, -1 slightly disagree, -3 moderately disagree, -5 strongly disagree)

a) Save the healthy dog     +5     +3     +1     -1     -3     -5

b) Save the injured man     +5     +3     +1     -1     -3     -5

Who would you save?

a) The healthy dog
b) The injured man

Why?

## References

1. Signorelli, C.M.: Can Computers become Conscious and overcome Humans? Front. Robot. Artif. Intell. 5, 121 (2018).
2. Turing, A..: Computing Machinery and Intelligence. Mind. 59, 433–460 (1950).
3. Silver, D., Schrittwieser, J., Simonyan, K., Antonoglou, I., Huang, A., Guez, A., Hubert, T., Baker, L., Lai, M., Bolton, A., Chen, Y., Lillicrap, T., Hui, F., Sifre, L.: Mastering the game of Go without human knowledge. Nature. 550, 354–359 (2017).
4. Bringsjord, S., Licato, J., Sundar, N., Rikhiya, G., Atriya, G.: Real Robots that Pass Human Tests of Self-Consciousness. In: Proceeding of the 24th IEEE International Symposium on Robot and Human Interactive Communication. pp. 498–504 (2015).
5. Legg, S., Hutter, M.: Universal intelligence: A definition of machine intelligence. Minds Mach. 17, 391–444 (2007).
6. Arsiwalla, X.D., Signorelli, C.M., Puigbo, J.-Y., Freire, I.T., Verschure, P.: What is the Physics of Intelligence? Front. Artif. Intell. Appl. Proceeding 21st Int. Conf. Catalan Assoc. Artif. Intell. 308, 283–286 (2018).
7. Arsiwalla, X.D., Sole, R., Moulin-Frier, C., Herreros, I., Sanchez-Fibla, M., Verschure, P.: The Morphospace of Consciousness. ArXiv. 20 (2017).
8. Coecke, B., Sadrzadeh, M., Clark, S.: Mathematical Foundations for a Compositional Distributional Model of Meaning. Linguist. Anal. 36, 345–384 (2010).
9. Bruza, P.D., Wang, Z., Busemeyer, J.R.: Quantum cognition: a new theoretical approach to psychology. Trends Cogn. Sci. 19, 383–93 (2015).
10. Kiverstein, J., Miller, M.: The embodied brain: towards a radical embodied cognitive neuroscience. Front. Hum. Neurosci. 9, (2015).
11. Arsiwalla, X.D., Verschure, P.: The global dynamical complexity of the human brain network. Appl. Netw. Sci. 1, 16 (2016).
12. Arsiwalla, X.D., Signorelli, C.M., Puigbo, J., Freire, I.T., Verschure, P.: Are Brains Computers, Emulators or Simulators? In: Springer (ed.) Conference on Biomimetic and Biohybrid Systems. pp. 11–15. , Paris (2018).
13. Arsiwalla, X.D., Herreros, I., Verschure, P.F.M.J.: On Three Categories of Conscious Machines. Biomim. Biohybrid Syst. Living Mach. 2016. Lect. Notes Comput. Sci. 9793, 389–392 (2016).
14. Gilovich, T., Griffin, D., Kahneman, D.: Heuristics and biases: The psychology of intuitive judgment. Cambridge University Press, Cambridge, UK (2002).
15. Pothos, E.M., Busemeyer, J.R.: Can quantum probability provide a new direction for cognitive modeling? Behav. Brain Sci. 36, 255–74 (2013).
16. Wang, Z., Solloway, T., Shiffrin, R.M., Busemeyer, J.R.: Context effects produced by question orders reveal quantum nature of human judgments. Proc. Natl. Acad. Sci. U. S. A. 111, 9431–6 (2014).
17. Aerts, D., Gabora, L., Sozzo, S.: Concepts and their dynamics: a quantum-theoretic modeling of human thought. Top. Cogn. Sci. 5, 737–72 (2013).
18. White, L.C., Barqué-Duran, A., Pothos, E.M.: An investigation of a quantum probability model for the constructive effect of affective evaluation. Philos. Trans. R. Soc. A Math. Phys. Eng. Sci. 374, (2016).
19. Petrinovich, L., O'Neill, P.: Influence of wording and framing effects on moral

intuitions. Ethol. Sociobiol. 17, 145–171 (1996).
20. Searle, J.R.: Minds, brains, and programs. Behav. Brain Sci. 3, 417–457 (1980).
21. Lindquist, K.A., Wager, T.D., Kober, H., Bliss-Moreau, E., Barrett, L.F.: The brain basis of emotion: A meta-analytic review. Behav. Brain Sci. 35, 121–143 (2012).
22. Hauser, M., Cushman, F.A., Young, L., Jin, R.K.X., Mikhail, J.: A dissociation between moral judgments and justifications. Mind Lang. 22, 1–21 (2007).
23. Christensen, J.F., Flexas, A., Calabrese, M., Gut, N.K., Gomila, A.: Moral judgment reloaded: A moral dilemma validation study. Front. Psychol. 5, 1–18 (2014).
24. Bekoff, M., Pierce, J.: Wild Justice: The Moral Lives of Animals. The University of Chicago Press (2009).
25. Greene, J.D., Sommerville, R.B., Nystrom, L.E., Darley, J.M., Cohen, J.D.: An fMRI Investigation of Emotional Engagement in Moral Judgment. Science (80-. ). 293, 2105–2108 (2001).
26. Moll, J., Zahn, R., Oliveira-Souza, R. de, Krueger, F., Grafman, J.: The neural basis of human moral cognition. Nat. Rev. Neurosci. 6, 799–809 (2005).
27. Grefenstette, E., Sadrzadeh, M.: Experimental Support for a Categorical Compositional Distributional Model of Meaning. In: Conference on Empirical Methods in Natural Language Processing. pp. 1394–1404. , Edinburgh (2011).
28. Coecke, B., Lewis, M.: A compositional explanation of the 'pet fish' phenomenon. In: Lecture Notes in Computer Science (including subseries Lecture Notes in Artificial Intelligence and Lecture Notes in Bioinformatics). pp. 179–192 (2016).
29. Bolt, J., Coecke, B., Genovese, F., Lewis, M., Marsden, D., Piedeleu, R.: Interacting Conceptual Spaces I : Grammatical Composition of Concepts. ArXiv. (2017).
30. Yearsley, J.M., Busemeyer, J.R.: Quantum cognition and decision theories: A tutorial. J. Math. Psychol. 74, 99–116 (2016).
31. Lambek, J.: From word to sentence. Polimetrica, Milan. (2008).
32. Gardenfors, P.: Conceptual Spaces as a Framework for Knowledge Representation. Mind Matter. 2, 9–27 (2004).
33. Aerts, D., Gabora, L., Sozzo, S., Veloz, T.: Quantum structure in cognition: fundamentals and applications. V. Privman V. Ovchinnikov (Eds.), IARIA, Proc. Fifth Int. Conf. Quantum, Nano Micro Technol. 57–62 (2011).
34. Veloz, T.: Toward a Quantum Theory of Cognition: History, Development, and Perspectives, (2016).
35. Varela, F.J.: Neurophenomenology: A Methodological Remedy for the Hard Problem. J. Conscious. Stud. 3, 330–349 (1996).
36. Mohammad, S.M., Turney, P.D.: Crowdsourcing a word-emotion association lexicon. Comput. Intell. 29, 436–465 (2013).
37. Signorelli, C.M.: Types of Cognition and Its Implications for Future High-Level Cognitive Machines. AAAI Spring Symp. Ser. SS-17-01, 622–627 (2017).